\documentclass[letterpaper, 10 pt, conference]{ieeeconf}  

\IEEEoverridecommandlockouts                              

\usepackage{xcolor}

\usepackage{graphicx}

\usepackage{gensymb}
\usepackage[hyphens]{url}
\usepackage{multirow}
\usepackage{multicol}
\usepackage{tabularx}
\usepackage{dsfont}
\usepackage{url}
\usepackage{color}
\usepackage{subcaption}
\usepackage{caption}
\usepackage{booktabs}
\usepackage{dcolumn}
\usepackage{amssymb}
\usepackage{hyperref}
\usepackage{amsmath}
\usepackage{colortbl}
\usepackage{tikz}
\colorlet{lightgray}{gray!20}

\usepackage{physics}
\usepackage{graphicx}
\usepackage{float}
\usepackage{subfloat}
\usepackage{longtable}
\usepackage{threeparttable}
\usepackage{xcolor}
\usepackage{diagbox}
\usepackage{cite}

\usepackage{color}
\hypersetup{
    colorlinks=false,
    linkcolor=blue,
    filecolor=magenta,      
    urlcolor=cyan,
    pdftitle={Overleaf Example},
    pdfpagemode=FullScreen,
    }

\usepackage{booktabs}
\usepackage{multirow}

 \renewcommand{\paragraph}[1]{
    \vspace{2mm}
     \noindent\textbf{#1} 
 }

\title{\LARGE \bf
S2R-ViT for Multi-Agent Cooperative Perception: Bridging the Gap from Simulation to Reality}

\author{Jinlong Li$^{1}$, Runsheng Xu$^{2}$, Xinyu Liu$^{1}$, Baolu Li$^{1}$, Qin Zou$^{3}$, Jiaqi Ma$^{2}$, Hongkai Yu$^{1*}$ 
\thanks{$^{1}$Cleveland State University,  Cleveland Vision $\&$ AI Lab.   $^{2}$University of California, Los Angeles,  UCLA Mobility Lab. $^{3}$Wuhan University, MVR Lab. }
\thanks{*Corresponding Author: h.yu19@csuohio.edu}
}

\begin{document}

 \maketitle
\thispagestyle{empty}
\pagestyle{empty}

\begin{abstract}
Due to the lack of enough real multi-agent data and time-consuming of labeling, existing multi-agent cooperative perception algorithms usually select the simulated sensor data for training and validating. However, the perception performance is degraded when these simulation-trained models are deployed to the real world, due to the significant domain gap between the simulated and real data. In this paper, we propose the \textit{first} Simulation-to-Reality transfer learning framework for multi-agent cooperative perception using a novel Vision Transformer, named as S2R-ViT, which considers both the Deployment Gap and Feature Gap between simulated and real data. We investigate the effects of these two types of domain gaps and propose a novel uncertainty-aware vision transformer to effectively relief the Deployment Gap and an agent-based feature adaptation module with inter-agent and ego-agent discriminators to reduce the Feature Gap. Our intensive experiments on the public multi-agent cooperative perception datasets OPV2V and V2V4Real demonstrate that the proposed S2R-ViT can effectively bridge the gap from simulation to reality and outperform other methods significantly for point cloud-based 3D object detection.





\end{abstract}

\section{Introduction}
The recent advancement in multi-agent cooperative perception shows potentials to overcome the limitation of single-agent perception that suffers from the challenges of perceiving range and occlusion~\cite{chen2019f,xu2023v2v4real}. By leveraging agent-to-agent communication technology to share information, multi-agent cooperative perception systems can significantly enhance perception performance compared to the single-agent perception~\cite{wang2020v2vnet,xu2022opv2v}. Due to the difficulties of collecting multi-agent data with communication in the real world, it is expensive and not easy to collect enough real data in diverse and complex real-world  environments~\cite{xu2023v2v4real}. Furthermore, the ground-truth data labeling and uniform coordinate projection for multi-agent cooperative perception systems is particularly time-consuming. Therefore, many existing multi-agent cooperative perception research works usually select the simulated data for model training and validating~\cite{wang2020v2vnet,xu2022opv2v}.

However, when we apply the models trained with simulated data to the real world, the perception performance is typically degraded. This phenomenon is because of the significant domain gap between the simulated and real data. In this paper, our research is focused on utilizing labeled simulated data and unlabeled real-world data as transfer learning to reduce the domain gap for multi-agent cooperative perception. We observe that the domain gap from simulation to reality for multi-agent cooperative perception includes the following two perspectives.

\begin{figure}[!t]
\centering
\subfloat{ 
\includegraphics[width=1\columnwidth]{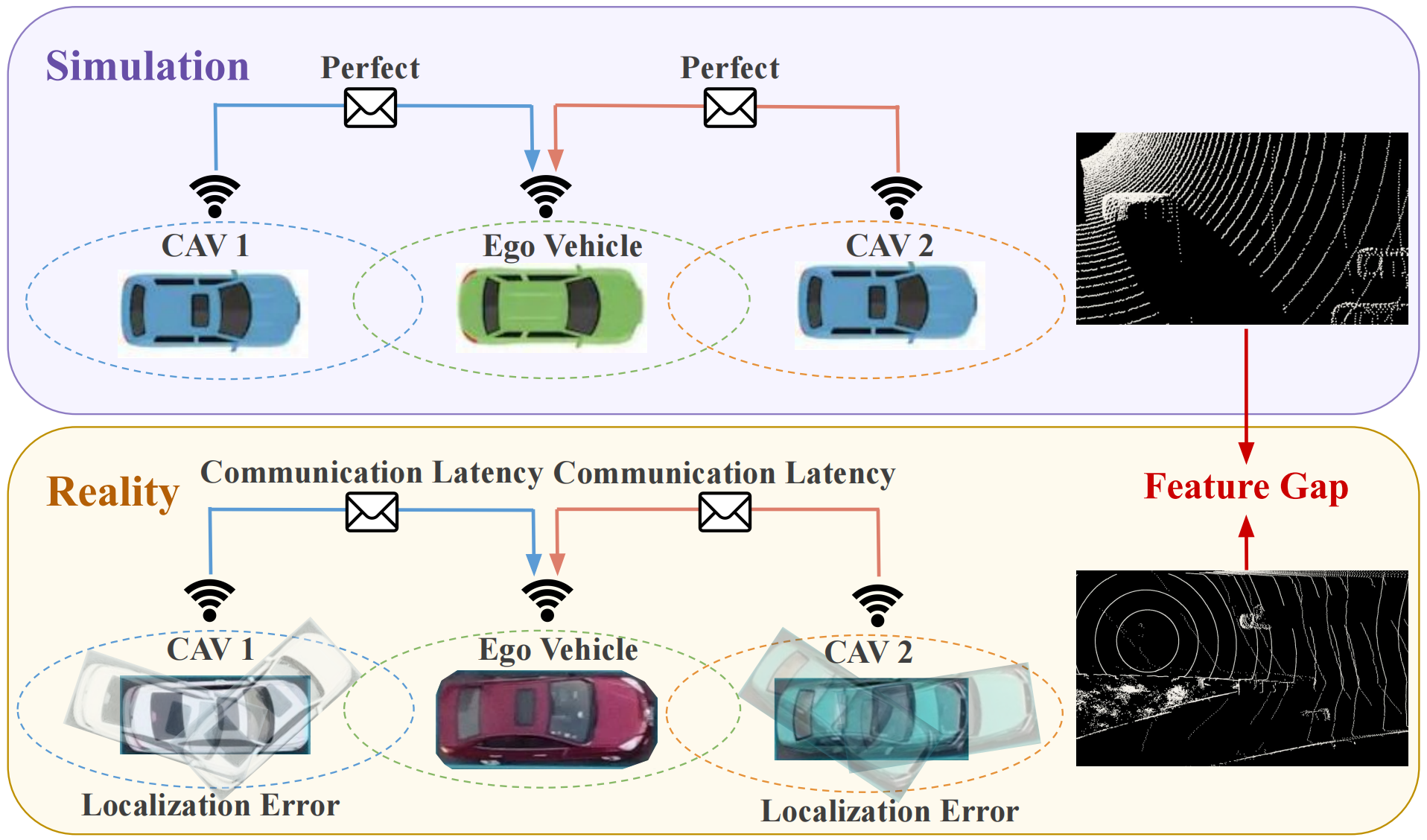}%
}
\caption{Illustration of the domain gap (\textit{Deployment Gap}, \textit{Feature Gap}) for multi-agent cooperative perception from simulation to reality. Here we use Vehicle-to-Vehicle (V2V) cooperative perception in autonomous driving as example. CAV indicates the Connected Autonomous Vehicles.}
\label{fig:motivation}
\end{figure}

\begin{itemize}
    \item \textbf{Deployment Gap:} As shown in Fig.~\ref{fig:motivation}, different with the ideal simulation setting, the multiple agents might have localization (positional and heading) errors due to the unavoidable GPS errors and communication latency (time delay) during the real-world agent-to-agent communication. 

    \item \textbf{Feature Gap:} As illustrated in  Fig.~\ref{fig:motivation}, the point cloud feature distribution in real world might differ significantly with that of the simulated  data, such as more complex driving scenarios, different LiDAR channel numbers, mixed traffic flow, various point cloud variations, and so on.    
\end{itemize}

\begin{figure*}[htb] 
    \begin{centering}
            \includegraphics[width=1\textwidth]{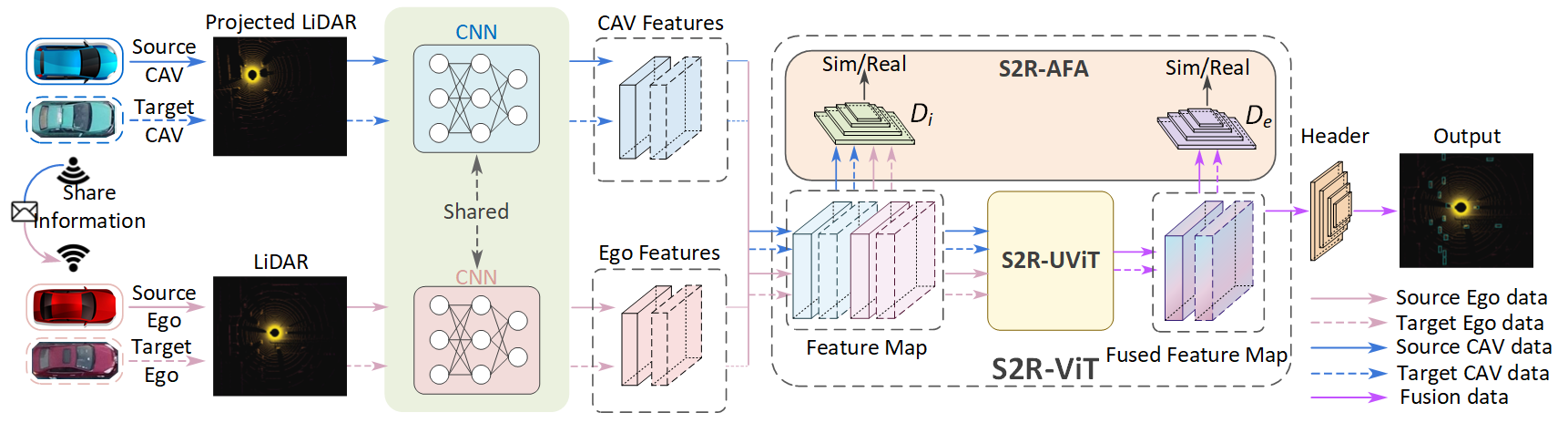}
    \par\end{centering}
    \caption{Overview of the proposed \textbf{S2R-ViT} for multi-agent cooperative perception from simulation to reality, which leverages \textit{S2R-UViT} module to handle uncertainty in Deployment Gap and tackles the Feature Gap through \textit{S2R-AFA} module including the inter-agent  discriminator $D_i$ and the ego-agent discriminator $D_e$. Source Domain: labeled simulated data, Target Domain: unlabeled real-world data. Best viewed in color.}
    \label{fig:architecture}
\end{figure*}


In this paper, we propose the first Simulation-to-Reality (S2R) transfer learning framework for multi-agent cooperative perception  using a novel Vision Transformer (ViT), named as S2R-ViT,  by taking both Deployment Gap and Feature Gap into consideration. We choose the task of Vehicle-to-Vehicle (V2V) Cooperative Perception for the point cloud-based 3D object detection as algorithm development. Specifically, our framework takes the labeled point cloud data from simulation and the unlabeled data from real world as input, so as to largely utilize the simulated data. In machine learning research, this setting is widely called as Unsupervised Domain Adaptation~\cite{song2020multi} from source domain (simulation) to target domain (reality).

The proposed S2R-ViT comprises two key components: (1) S2R-UViT: a novel S2R Uncertainty-aware Vision Transformer to effectively relief the \textit{uncertainties brought by the Deployment Gap}. Specifically, S2R-UViT includes a Local-and-Global Multi-head Self Attention (LG-MSA) module to enhance feature interactions across all agents' spatial positions to tolerate the uncertainty drawbacks and also a Uncertainty-Aware Module (UAM) to enhance the ego-agent features by considering the shared other-agent features of different uncertainty levels. (2) S2R-AFA: S2R Agent-based Feature Adaptation to reduce the \textit{Feature Gap}. S2R-AFA utilizes the inter-agent and ego-agent discriminators to extract domain-invariant features to bridge the Feature Gap. Finally, we conducted extensive experiments on two public datasets, namely simulated OPV2V~\cite{xu2022opv2v} and real V2V4Real~\cite{xu2023v2v4real}, to justify the effectiveness of our proposed method. Our contributions are summarized as follows.

\begin{itemize}
    \item To the best of our knowledge, we propose the \textbf{first research}, named S2R-ViT, on multi-agent cooperative perception from simulation to reality by  investigating two types of domain gaps,  \textit{i.e.}, Deployment Gap and Feature Gap, for point cloud-based 3D object detection. 

    \item We propose a novel Uncertainty-aware Vision Transformer (S2R-UViT) to effectively relieve the uncertainties brought by the Deployment Gap.

    \item We design an Agent-based Feature Adaptation (S2R-AFA) module that includes inter-agent and ego-agent discriminators to effectively reduce the Feature Gap between simulation and reality.

    \item We evaluate our proposed method on the large-scale simulated OPV2V dataset and the real V2V4Real dataset, whose experiments demonstrate our superior performance in point cloud-based 3D object detection.    
\end{itemize} 


\begin{figure*}[htb] 
    \begin{centering}
        \includegraphics[width=0.79\textwidth]{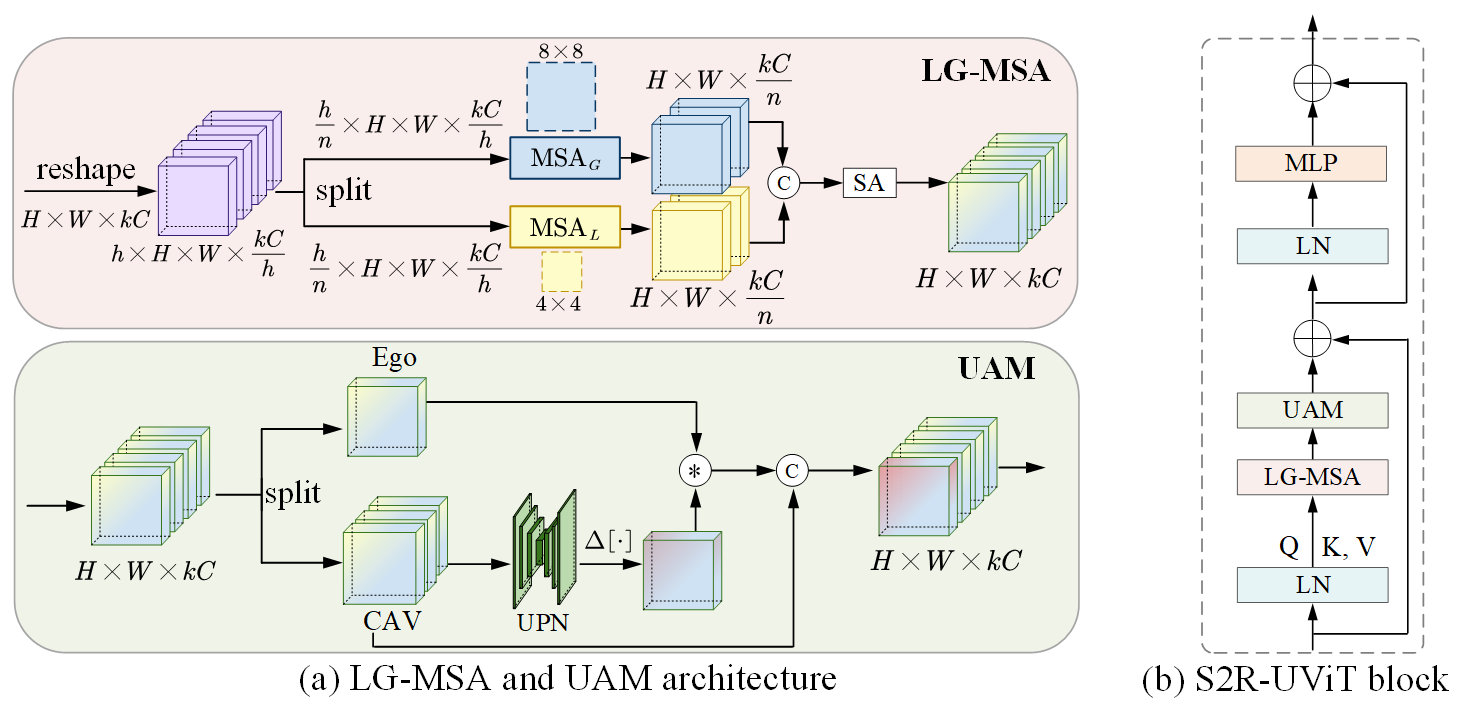}
    \par\end{centering}
    \caption{Architecture of the proposed S2R-UViT: Simulation-to-Reality Uncertainty-aware Vision Transformer.}
    \label{fig:S2R-ViT}
\end{figure*}

\section{Related Work}\label{Sec:Related_Work}

\noindent \textbf{Multi-Agent Perception.}
Multi-agent perception system can overcome the challenges of occlusion and short-range perceiving via agent-to-agent communication technology to achieve the large-range perceiving, which has attracted the attention of many researchers. Rather than exchanging raw sensing data or detected outputs, contemporary approaches typically share intermediate features extracted by neural networks. This strategy offers an optimal balance between accuracy and bandwidth requirements~\cite{xu2022v2x,xu2022bridging,huwhere2comm,chen2019f}.
V2VNet~\cite{wang2020v2vnet} utilizes a graph neural network to aggregate features from individual agents.
When2com~\cite{huwhere2comm} employs a spatial confidence-aware communication strategy aimed at enhancing performance by reducing communication overhead.
OPV2V~\cite{xu2022opv2v} proposes an attention fusion framework for cooperative perception system. 
SyncNet~\cite{lei2022latency} introduces a latency compensation module for time-domain synchronization. V2X-ViT~\cite{xu2022v2x} and CoBEVT~\cite{xu2022cobevt} propose transformer or axial attention based methods  to improve the performance. Although these methods have demonstrated impressive performance, all of them are implemented in the simulated data, while this paper aims to bridge the gap from simulation to reality.

\noindent \textbf{Challenges in Multi-Agent Perception.}
Multi-Agent perception system also introduces some new challenges, \textit{e.g.}, localization error, communication latency, adversarial attacks. These challenges might diminish the benefits of collaborations~\cite{xu2022bridging,10077757,xu2022v2x}. To ensure robustness for Multi-Agent perception, 
V2X-ViT~\cite{xu2022v2x}   proposes a ViT based multi-agent perception system, showcasing robust performance under communication delay. \cite{vadivelu2021learning} introduces a pose regression module and consistency module prior to feature aggregation to rectify pose errors. \cite{huwhere2comm} presents the first latency-aware multi-agent perception framework, achieving feature-level synchronization. \cite{tu2021adversarial} explores adversarial attacks in multi-agent perception and devises a novel transfer attack approach within the intermediate collaborative perception. \cite{10077757} proposes a lossy communication-aware repair network to ensure the robustness of multi-agent perception amidst lossy communication data. 
To promote multi-agent perception research, several large-scale cooperative perception datasets are publicized, \textit{e.g.}, simulation based OPV2V~\cite{xu2022opv2v} and V2XSet~\cite{xu2022v2x}, real-world V2V4Real~\cite{xu2023v2v4real}. Because collecting the multi-agent sensing data in the wild and annotating its ground truth are time and labor consuming, many existing  research works are based on simulated data. In this paper, we focus on improving point cloud-based cooperative 3D object detection in real world by largely utilizing labelled simulated data and just unlabelled real-world data.

\noindent \textbf{Domain Adaptation for Perception.} 
Domain adaptation involves adapting the machine learning model trained on source domain to the target domain. Many domain adaptation works are mainly focused on the RGB camera data~\cite{li2021domain,li2023domain,oza2023unsupervised,hoyer2023mic}, while more domain adaptation works have been proposed to solve this problem in LiDAR data~\cite{yi2021complete,xu2021spg, saltori2022cosmix,xu2022bridging}. Specifically,
Semantic Point Generation (SPG)~\cite{xu2021spg} is presented to enhance the reliability of LiDAR detectors against domain gaps to generates semantic points at the 3D objects.
CoSMix~\cite{saltori2022cosmix} is designed for 3D LiDAR segmentation, aiming to alleviate the domain gap by generating two intermediate domains comprising composite point clouds. This is achieved through the application of a special  mixing strategy at the input level. 
\cite{xu2022bridging} proposes a domain adaption framework to bridge the domain gap of shared data in communication for collaborative perception. In this paper, an agent-based feature adaptation module are proposed to reduce the feature gap between simulated to real data for multi-agent perception.

\section{Methodology}\label{Sec:Method}


As illustrated in Fig.~\ref{fig:architecture}, the proposed S2R work is an end-to-end unified deep learning pipeline,  including 1) V2V metadata sharing, 2) Feature extraction and sharing, 3) S2R-ViT, and 4) Detection header.

\subsection{Overview of Architecture}\label{}
\noindent \textbf{1) V2V metadata sharing.}
We designate one of the CAVs as the ego vehicle and construct a spatial graph around it, with each node representing a CAV within the communication range. After receiving the ego vehicle's relative pose and extrinsic information, all other nearby CAVs then project their LiDAR data to the coordinate frame of the ego vehicle. 
\noindent \textbf{2) Feature extraction and sharing.}
We utilize the PointPillar backbone~\cite{2019PointPillars} to extract intermediate features from LiDAR data due to its low inference latency and refined memory usage. Each CAV possesses its own feature extraction module. Following feature extraction by each CAV, the ego vehicle receives the visual features of neighboring CAVs via communication. 
\noindent \textbf{3) S2R-ViT.} The intermediate features aggregated from other surrounding CAVs are fed into our  major component named S2R-ViT, which consists of S2R-UViT and S2R-AFA modules. 
Upon receiving the final fused features, we utilize a prediction header for 3D object classification and localization.

\subsection{S2R-UViT: Simulation-to-Reality Uncertainty-aware Vision Transformer}\label{sim2real-vit}
The Deployment Gap from simulation to reality brings different uncertainties to both ego and neighboring agents, \textit{e.g.}, spatial bias by GPS errors, spatial misalignment in the coordinate projection because of communication latency. \textit{How to effectively reduce the degradation effects of these uncertainty drawbacks is an essential problem and open question to the S2R multi-agent perception research.} In this paper, we propose to answer this question from two perspectives: uncertainties can be relieved by enhancing (1) the feature interactions across all agents' spatial positions more comprehensively and (2) the ego-agent features by considering the shared other-agent features of different uncertainty levels. These two perspectives motivate us to develop the novel Local-and-Global Multi-head Self Attention (LG-MSA) Module and Uncertainty-Aware Module (UAM) respectively. The Fig.~\ref{fig:S2R-ViT} presents the two major modules of the proposed S2R-UViT.

\subsubsection{Local-and-Global Multi-head Self Attention (LG-MSA)} 
In order to enhance the feature interactions across all agents' spatial positions more comprehensively, we propose LG-MSA to promote both local and global feature interactions across all agents' spatial positions. In the proposed LG-MSA, a local feature-based attention is utilized to focus on the local details of spatial features, while a global feature-based attention is used to pay attention on the wide range of spatial features. Specifically, the fused features of the ego agent and other agents are spilt into two branches (\textit{i.e.}, local branch and global branch) to process local-based and global-based attentions, respectively, then these two features are concatenated together and fed into a Self Attention (SA) module to further capture local and global information.

Inspired by~\cite{liu2021swin,ren2022beyond}, after retrieving the whole features $F_{e,o} \in \mathbb{R}^{H \times W \times kC}$ with spatial dimension $H \times W $ from all agents ($e$: ego-agent, $o$:  other agents, $k$: number of all agents), we reshape it into $\mathbb{R}^{ h \times H \times W \times \frac{kC}{h}}$. After that, we take $h = 8$ as head number and $n = 2$ as window-type number, the multi-head $h$ of standard multi-head Self Attention Module (MSA)~\cite{dosovitskiy2021an} is evenly divided into two groups with different window sizes, \textit{i.e.}, $4 \times 4 $ for the local branch and $8 \times 8$ for the global branch. In the \textit{local branch}, the spilt feature $F_{e,o}^{l} \in \mathbb{R}^{ \frac{h}{n} \times H \times W \times \frac{kC}{h}}$ is fed into the MSA$_L$ with a small window size $4 \times 4$ to enhance the local details of spatial features. In the \textit{global branch}, another spilt feature $F_{e,o}^{g} \in \mathbb{R}^{ \frac{h}{n} \times H \times W \times \frac{kC}  {h}}$ is fed into the MSA$_G$ with a large window size $8 \times 8$ to capture global spatial feature information. Then we concatenate these two-branch output features and implement the feature interactions by a  Self Attention (SA) module to obtain the promoted feature $F_{e,o}^{p} \in \mathbb{R}^{H \times W \times kC}$. As show in Fig.~\ref{fig:S2R-ViT}(a), the proposed LG-MSA computation can be defined as

\begin{equation}
    F_{e,o}^{p} = \mathrm{SA}(\mathrm{Concat}(\mathrm{MSA}_{L}(F_{e,o}^{l}), \mathrm{MSA}_{G}(F_{e,o}^{g}))).
    \label{eq:SA}
\end{equation}


\subsubsection{Uncertainty-Aware Module (UAM)} 
To enhance the ego-agent feature based on the shared other-agent features, their different uncertainty levels should not be neglected. When receiving the feature $F_{e,o}^{p}$ for all agents, we split them into ego-agent feature $F_{e}^{p}  \in \mathbb{R}^{H \times W \times C} $ and other agent shared feature $F_{o}^{p}  \in \mathbb{R}^{H \times W \times (k-1)C}$. The shared other-agent feature $F_{o}^{p}$ will be fed into an Uncertainty Prediction Network (UPN) to predict the uncertainty levels on spatial features, which generates a uncertainty-level map $M  \in \mathbb{R}^{H \times W \times (k-1)C}$ with the same spatial size of input. The UPN is an encoder-decoder based neural network simplified from~\cite{mildenhall2018burst}, which can be learned during the end-to-end training of our  whole architecture. Inspired by the natural selection in the  mechanism of evolution, the feature values in the predicted uncertainty-level map $M$ with high uncertainty levels (\textit{i.e.}, low confidences) are reset as 1 using the median as threshold. It results in a new uncertainty-level map $M_t$. Then, $M_t$ is multiplied with $F_{e}^{p}$, by taking the shared other-agent features of different uncertainty levels into consideration, so as to produce the enhanced ego feature. In other words, only the low-uncertainty spatial features of other agents (related to non-ones in $M_t$) will contribute to enhance the ego-agent feature during the matrix multiplication. Finally, the enhanced ego feature and  other agent shared feature are concatenated together to obtain the combined feature $F_{e,o}^{h} \in \mathbb{R}^{H \times W \times kC}$. As show in Fig.~\ref{fig:S2R-ViT}(a), the computation of the proposed UAM can be formulated as

 
\begin{equation}
    F_{e,o}^{h} = \mathrm{Concat}(\Delta[\mathrm{UPN}(F_{o}^{p})] \circledast F_{e}^{p},F_{o}^{p}),   
    \label{eq:output}
\end{equation}
where $\Delta[\cdot]$ represents the threshold process and $\circledast$ denotes the matrix dot product. Combining these local and global attention branches with standard Transformer designs~\cite{liu2021swin,ren2022beyond}, such as Layer Normalization (LN), MLPs, and skip-connections, results in our proposed S2R-UViT block, as depicted in Fig.~\ref{fig:S2R-ViT}(b). The S2R-UViT block can be  expressed as:

\begin{equation}
    F_{e,o}^{h} = \mathrm{S2RAttn}(\mathrm{LN}(F_{e,o})) + F_{e,o},
    \label{eq:S2R}
\end{equation}
\begin{equation}
    \hat{F_{e,o}^{h}} = \mathrm{MLP}(\mathrm{LN}(F_{e,o}^{h})) + F_{e,o}^{h},
    \label{eq:block}
\end{equation}
where $F_{e,o}^{h}$ and $\hat{F_{e,o}^{h}}$ denote the output features of our S2RAttn module (\textit{i.e.}, the proposed LG-MSA and UAM) and MLP module.

\subsection{S2R-AFA: Simulation-to-Reality Agent-based Feature Adaptation}\label{ADA} 
To reduce the Feature Gap from simulated feature $F_s$ to real feature $F_r$ within both inter-agent and ego-agent features, we design two domain discriminators/classifiers before and after fusion as shown in Fig.~\ref{fig:architecture}, where all agent feature $F_s$ and $F_r$ before fusion are fed into the inter-agent discriminator $D_i$ to classify whether it belongs to simulation or reality, and the fused ego features $F_s^{e}$ and $F_r^{e}$ after fusion are also  classified into simulation or reality by the ego-agent discriminator $D_e$. The binary cross entropy loss is used for binary domain (Sim/Real) classification. These two discriminators are adversarially optimized by the Agent-based Feature Adaptation loss $\mathcal{L}_{AFA}$: 
\begin{equation}
    \min\limits_{G_m}\max\limits_{D_i, D_e}  \mathcal{L}_{AFA} = \mathbb{E}_{s,r}[D_i(F_s,F_r)] + \mathbb{E}_{s,r} [D_{e}(F_s^{e},F_r^{e})],
    \label{eq:GAN}
\end{equation}
where $\mathbb{E}_{s,r}$ indicates the domain classification error in simulation and reality 
respectively, $G_m$ is our whole model (backbone, S2R-UViT, and detection header) which can be thought as the  generator of Generative Adversarial  Network~\cite{radford2015unsupervised}. Because of S2R-AFA, our generator model $G_m$ will have the capability of extracting domain-invariant features of simulation and reality.

We use the focal loss~\cite{lin2017focal} and  smooth $L_1$ loss as the detection loss for our model training. Our final loss is the combination of detection loss and the Agent-based Feature Adaptation loss as follows,
\begin{equation}
    \mathcal{L}_{total} = w_1 \mathcal{L}_{det} + w_2 \mathcal{L}_{AFA},
   \label{eq:total_loss}
\end{equation}
where $w_1$ and $w_2$ are balance weights with sum as 1.




\begin{table}[]

\centering
\caption{3D detection performance on two OPV2V testing sets under \textbf{\textit{Deployment-Gap Scenario}} with \textit{Perfect Setting} and \textit{Noisy Setting}. All methods are trained on \textit{Perfect Setting}.}
\resizebox{1\columnwidth}{!}{%
\begin{tabular}{@{}cccccc@{}}
\toprule
\multirow{2}{*}{Models} & \multirow{2}{*}{Setting} & \multicolumn{2}{c}{V2V CARLA Towns} & \multicolumn{2}{c}{V2V  Culver City} \\
                          &                              & AP@0.5 & AP@0.7                     & AP@0.5 & AP@0.7 \\ \midrule
\multirow{2}{*}{AttFuse~\cite{xu2022opv2v}}  & \multicolumn{1}{c|}{Perfect} & 0.921  & \multicolumn{1}{c|}{0.804} & 0.887  & 0.716  \\
                          & \multicolumn{1}{c|}{Noisy}   & 0.851  & \multicolumn{1}{c|}{0.472} & 0.865  & 0.565  \\ \midrule
\multirow{2}{*}{V2VNet~\cite{wang2020v2vnet}}   & \multicolumn{1}{c|}{Perfect} & 0.915  & \multicolumn{1}{c|}{0.828} & 0.884  & 0.757  \\
                          & \multicolumn{1}{c|}{Noisy}   & 0.845  & \multicolumn{1}{c|}{0.413} & 0.868  & 0.617  \\ \midrule
\multirow{2}{*}{F-cooper~\cite{chen2019f}} & \multicolumn{1}{c|}{Perfect} & 0.907  & \multicolumn{1}{c|}{0.810}  & 0.893  & 0.746  \\
                          & \multicolumn{1}{c|}{Noisy}   & 0.842  & \multicolumn{1}{c|}{0.479} & 0.881  & 0.624  \\ \midrule
\multirow{2}{*}{V2X-ViT~\cite{xu2022v2x}}  & \multicolumn{1}{c|}{Perfect} & 0.902  & \multicolumn{1}{c|}{0.792} & 0.903  & 0.764  \\
                          & \multicolumn{1}{c|}{Noisy}   & 0.801  & \multicolumn{1}{c|}{0.395} & 0.882  & 0.606  \\ \midrule
\multirow{2}{*}{CoBEVT~\cite{xu2022cobevt}}   & \multicolumn{1}{c|}{Perfect} & 0.925  & \multicolumn{1}{c|}{0.852} & 0.904  & 0.776  \\
                          & \multicolumn{1}{c|}{Noisy}   & 0.862  & \multicolumn{1}{c|}{0.519} & 0.889  & 0.665  \\ \midrule
\multirow{2}{*}{V2VAM~\cite{10077757}}    & \multicolumn{1}{c|}{Perfect} & 0.916  & \multicolumn{1}{c|}{0.849} & 0.903  & \textbf{\color{cyan}0.794}  \\
                          & \multicolumn{1}{c|}{Noisy}   & 0.876  & \multicolumn{1}{c|}{0.507} & 0.883  & 0.663  \\ \midrule
\multirow{2}{*}{S2R-UViT}    & \multicolumn{1}{c|}{Perfect} & \textbf{\color{cyan}0.928} & \multicolumn{1}{c|}{\textbf{\color{cyan}0.867}} & \textbf{\color{cyan}0.912}  & 0.768     \\
                          & \multicolumn{1}{c|}{Noisy}   & \textbf{0.879} & \multicolumn{1}{c|}{\textbf{0.579}} & \textbf{0.900} & \textbf{0.681}     \\ \bottomrule
\end{tabular}%
}
\label{tab-Development}
\end{table}

\begin{figure*}[htb] 
    \begin{centering}
        \includegraphics[width=0.97\textwidth]{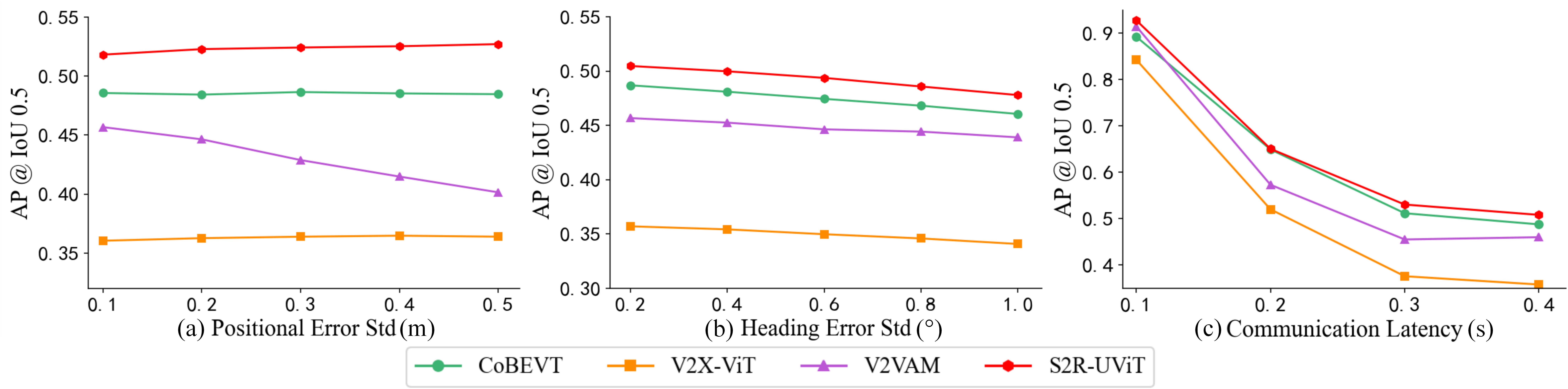}
    \par\end{centering}
    \caption{Robustness in  \textbf{\textit{Deployment-Gap Scenario}} including GPS (positional, heading) errors  and communication latency on the CARLA Towns testing set of OPV2V~\cite{xu2022opv2v} with the simulated \textit{Noisy Setting}.}
    \label{fig:development-gap-reasult}
\end{figure*}

\section{Experiment}\label{Sec:Experiment}

\subsection{Dataset}
We conduct experiments on two public benchmark datasets (OPV2V~\cite{xu2022opv2v},  V2V4Real~\cite{xu2023v2v4real}) for the V2V cooperative perception task.  
\textbf{OPV2V} is a large-scale \textit{simulated} dataset for V2V cooperative perception, which is collected by CARLA simulator and OpenCDA~\cite{xu2021opencda}. It contains 73 divergent scenes with various numbers of connected vehicles ($[1,5]$), and its training/validation/testing set is split into 6,764, 1,981, and 2,719 frames, respectively. 
\textbf{V2V4Real} is a \textit{real-world}, large-scale dataset with diverse driving scenarios, which is collected by two CAVs driving simultaneously in Columbus, Ohio, USA. It's split into the train/validation/test set with 14,210/2,000/3,986 frames, respectively.

\subsection{Experiments Setup}
\noindent \textbf{Evaluation Metrics.} 
We assess the perception performance based on the final 3D vehicle detection accuracy. Following~\cite{xu2022opv2v,xu2022v2x}, we define the evaluation range as $x\in[-140, 140]$ meters and $y\in[-40, 40]$ meters, encompassing all CAVs within this spatial range in the experiment. Accuracy is calculated as Average Precision (AP) at two Intersection-over-Union (IoU) thresholds ($0.5$ and $0.7$).

\noindent \textbf{Experiment Settings.}
In this work, we want to address the deployment gap and feature gap on LiDAR-based object detection and assess our model under two distinct settings:
\begin{itemize}
    \item[1)] \textbf{\textit{Deployment-Gap Scenario}}: During the training, all models are trained in the perfect simulated OPV2V training set, then all of them are evaluated on OPV2V CARLA Towns and Culver City testing sets under two different setting (\textit{e.g.}, \textit{Perfect} and \textit{Noisy}), respectively. 
    We implement the \textit{Noisy Setting} following~\cite{xu2022v2x} to simulate the GPS errors,
    the positional and heading noises of the transmitter are sampled from a Gaussian distribution with a standard deviation of $0.2$ m and $0.2^{\circ}$ respectively, and the communication latency is set as $100$ ms for all the evaluated models. This Deployment-Gap Scenario only includes the Deployment Gap.

    \item[2)] \textbf{\textit{Sim2Real  Scenario}}: We set the labeled training set of simulation dataset  OPV2V~\cite{xu2022opv2v} as the source domain and the unlabeled training set of the real-world dataset  V2V4Real~\cite{xu2023v2v4real} as the target domain for all models during the training, by following the same setting in~\cite{xu2023v2v4real}. Then, all trained models are evaluated on the testing set of V2V4Real to report the performance. This Sim2Real Scenario includes both the  Deployment Gap and Feature Gap from simulation to reality. 

\end{itemize} 

All methods utilize PointPillar~\cite{lang2019pointpillars} as the point cloud encoder/backbone. We employ the Adam optimizer~\cite{loshchilov2017decoupled} with an initial learning rate of $10^{-3}$ and gradually decay it every $10$ epochs based a factor of $0.1$. Following the hyperparameters outlined in V2X-ViT~\cite{xu2022v2x}, we train all models on two RTX 3090 GPU cards.

\noindent \textbf{Compared Methods.} Six state-of-the-art V2V methods are evaluated, all of which employ \textit{Intermediate Fusion} as the primary strategy: 
AttFuse~\cite{xu2022opv2v}, V2VNet~\cite{wang2020v2vnet}, 
F-Cooper~\cite{chen2019f}, V2X-ViT~\cite{xu2022v2x}, CoBEVT~\cite{xu2022cobevt}, and V2VAM~\cite{10077757}. We first train these methods on the perfect setting of OPV2V training set, and then these methods are evaluated on the  \textit{Noisy Setting} of OPV2V testing set and V2V4Real testing set to assess their performance. In addition, to show the effectiveness of reducing Feature Gap, two domain adaptation methods \textit{i.e.}, gradient reverse layer (GRL)~\cite{ganin2015unsupervised}  and adversarial gradient reverse layer (AdvGRL)~\cite{li2023domain} are utilized to  
back-propagate the reversed gradient to adversarially guide the model for generating domain-invariant features by one feature-level domain classifier (after fusion) and one object/proposal-level domain classifier (in detection header).


\begin{table}[htb]
\centering
\tiny
\caption{3D detection performance on V2V4Real testing set under \textbf{\textit{Sim2Real  Scenario}}. All methods with domain adaptation are trained following the setting in \textbf{\textit{Sim2Real Scenario}}. S2R-UViT w/ S2R-AFA indicates our S2R-ViT.}
\resizebox{0.8\columnwidth}{!}{%
\begin{tabular}{@{}lcc@{}}
\toprule
\multirow{2}{*}{Method} & \multicolumn{2}{c}{V2V4Real Testing} \\
                                  & AP@0.5           & AP@0.7          \\ \midrule
AttFuse~\cite{xu2022opv2v}        & 0.225           & 0.094          \\
AttFuse w/ GRL           & 0.356           & 0.139          \\
AttFuse w/ AdvGRL        & 0.366           & 0.137          \\ \midrule
V2VNet~\cite{wang2020v2vnet}      & 0.268           & 0.108          \\
V2VNet w/ GRL            & 0.376           & 0.122          \\
V2VNet w/ AdvGRL         & 0.358           & 0.103          \\ \midrule
F-Cooper~\cite{chen2019f}  & 0.236           & 0.091          \\
F-Cooper w/ GRL          & 0.372           & 0.116          \\
F-Cooper w/ AdvGRL       & 0.364           & 0.135          \\ \midrule
V2X-ViT~\cite{xu2022v2x}   & 0.274           & 0.103          \\
V2X-ViT w/ GRL           & 0.395           & 0.156          \\
V2X-ViT w/ AdvGRL        & 0.404           & 0.157          \\ \midrule
CoBEVT~\cite{xu2022cobevt}    & 0.334           & 0.130           \\
CoBEVT w/ GRL            & 0.398           & 0.163          \\
CoBEVT w/ AdvGRL         & 0.393           & 0.163          \\ \midrule
V2VAM~\cite{10077757}    & 0.332           & 0.120           \\
V2VAM w/ GRL             & 0.390           & 0.146          \\
V2VAM w/ AdvGRL          & 0.401           & 0.161          \\ \midrule
S2R-UViT                 & \textbf{\color{cyan}0.367}           & \textbf{\color{cyan}0.138}             \\
S2R-UViT w/ GRL      & 0.414           & 0.141             \\ 
S2R-UViT w/ AdvGRL    & 0.414           & 0.157             \\ 
S2R-UViT w/ S2R-AFA   &\textbf{0.441}  & \textbf{0.170}             \\ \bottomrule
\end{tabular}%
}
\label{tab:domaingap}
\vspace{-1.5em}
\end{table}

\begin{figure*}[thbp]%
  \centering%
 
  \subfloat[V2X-ViT w/ AdvGRL ]{%
    \resizebox{0.33\linewidth}{!}{
      \begin{tikzpicture}%
        \node at(0.0,0.0){\fbox{\includegraphics[width=\linewidth]{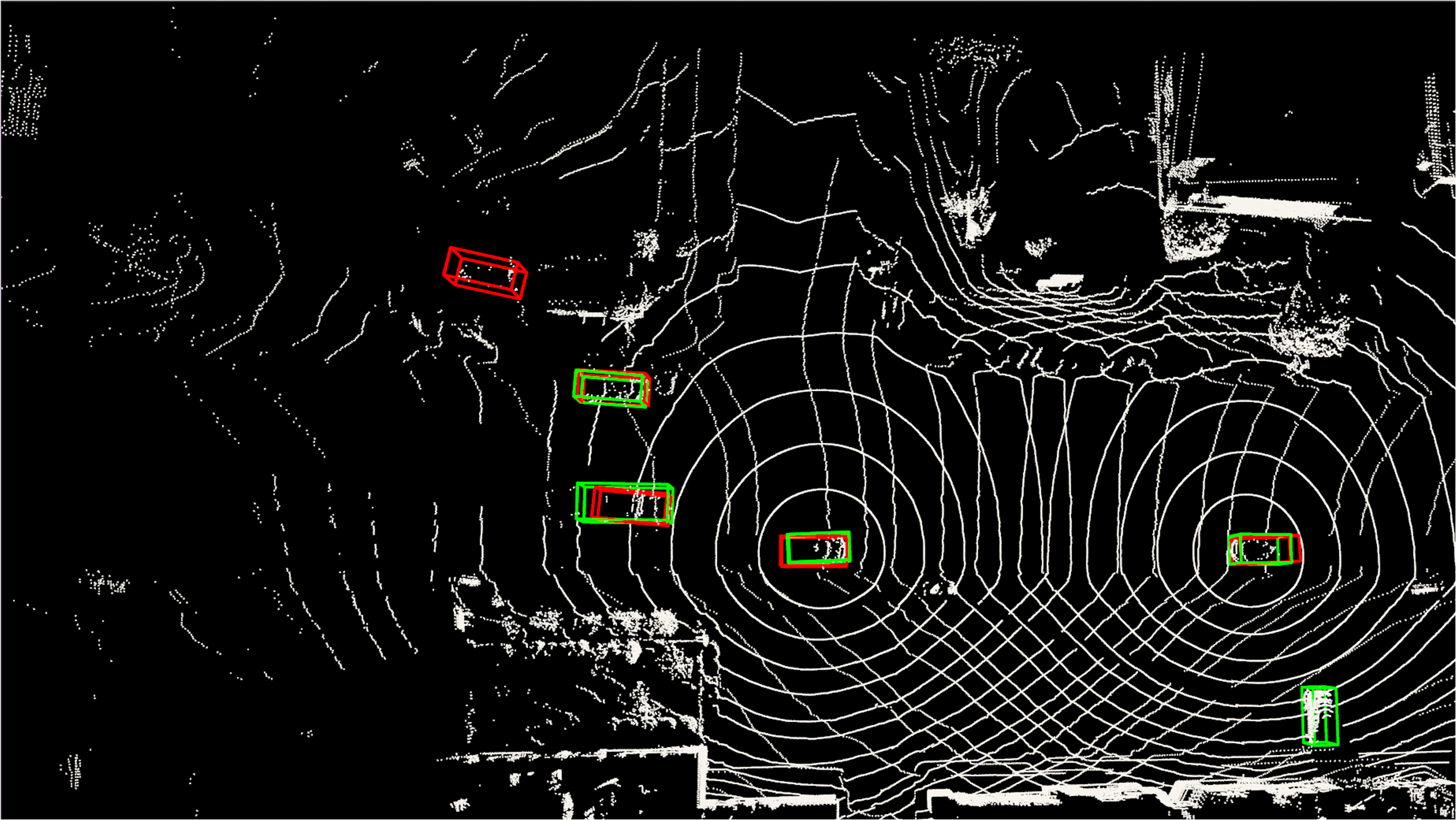}}};%
      \end{tikzpicture}%
    }%
  }%
  \subfloat[V2VAM w/ AdvGRL ]{%
    \resizebox{0.33\linewidth}{!}{
      \begin{tikzpicture}%
        \node at(0.0,0.0){\fbox{\includegraphics[width=\linewidth]{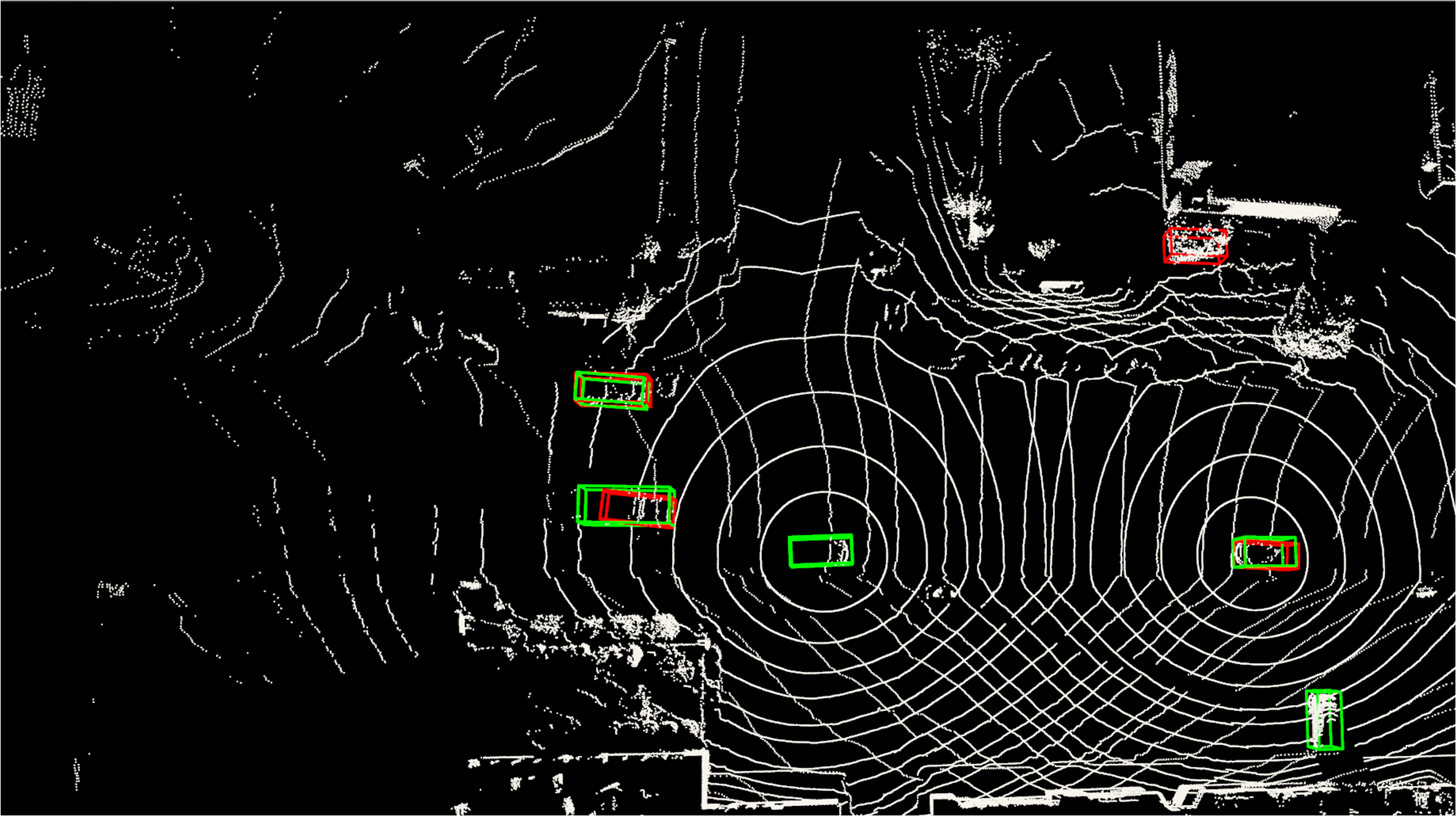}}};%
      \end{tikzpicture}%
    }%
  }%
  \subfloat[S2R-ViT]{%
    \resizebox{0.33\linewidth}{!}{
      \begin{tikzpicture}%
        \node at(0.0,0.0){\fbox{\includegraphics[width=\linewidth]{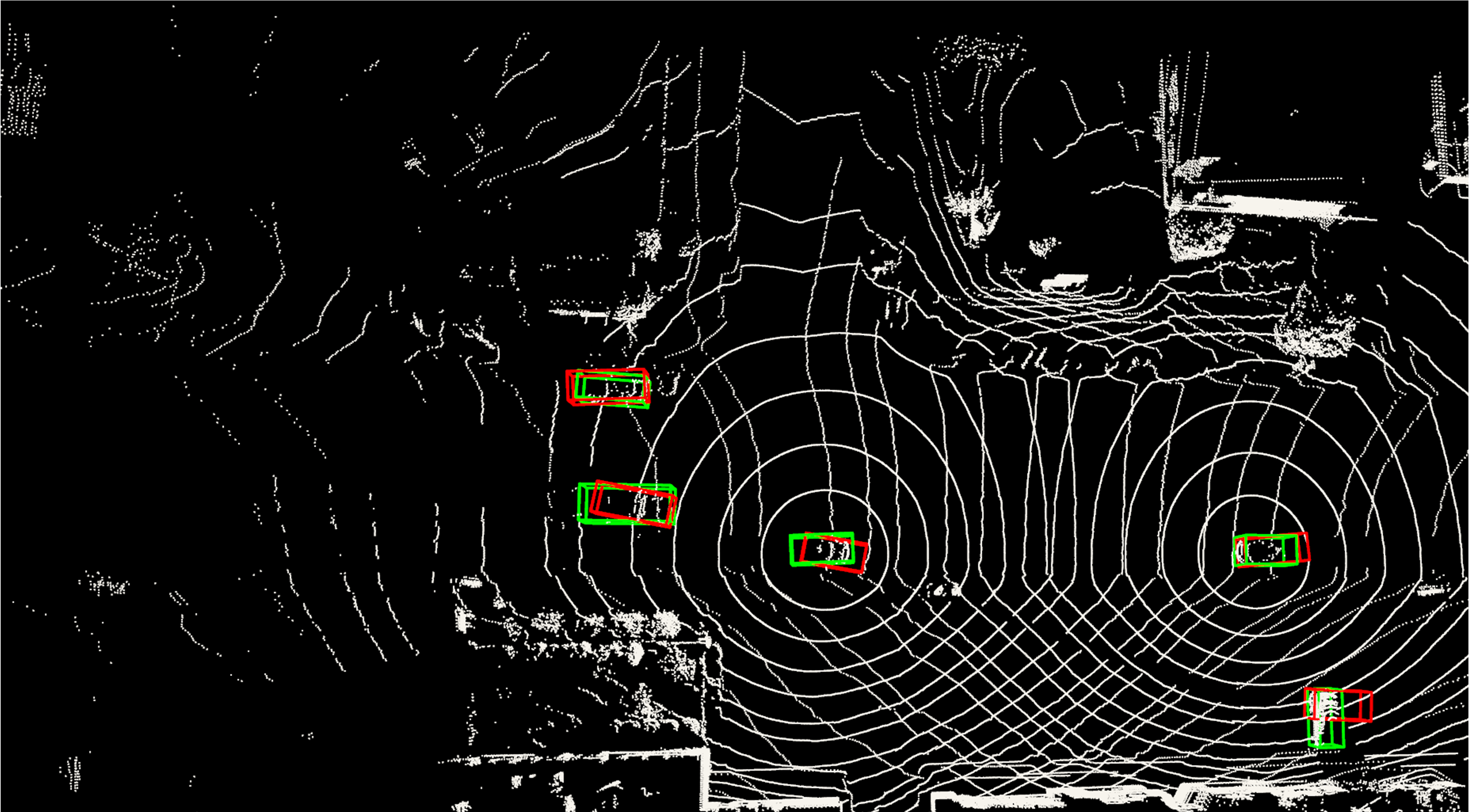}}};%
      \end{tikzpicture}%
    }%
  }%
  \caption{Visualization example of point cloud-based 3D object detection on V2V4Real testing set under the \textbf{\textit{Sim2Real  Scenario}}. \textcolor{green}{Green} and \textcolor{red}{red} 3D bounding boxes represent \textcolor{green}{ground truth} and \textcolor{red}{prediction} respectively. Best viewed in color.}%
  \vspace{-1.5em}
  \label{fig:detect}%
\end{figure*}


\subsection{Quantitative Evaluation}

\noindent \textbf{Performance in Deployment-Gap Scenario.}
Table~\ref{tab-Development} shows the performance comparison on \textbf{\textit{Deployment-Gap Scenario}}, where all methods are evaluated on \textit{Perfect Setting} and \textit{Noisy setting}, respectively. Under the \textit{Perfect Setting}, all cooperative perception methods achieve outstanding performance. Nevertheless, when these methods deployed on the \textit{Noisy Setting} scenario, which have the deployment gap with the \textit{Perfect Setting} scenario, the V2X-ViT~\cite{xu2022v2x}, CoBEVT~\cite{xu2022cobevt}, and V2VAM~\cite{10077757} drop $39.7\%$, $33.3\%$, and $34.2\%$ on for AP$@0.7$ on  V2V CARLA Towns testing set. It indicates the highly negative impacts by V2V deployment gap. While our proposed S2R-UViT achieves the mostly best performance under both \textit{Perfect Setting} and \textit{Noisy Setting} scenarios, which is highlighted in Table~\ref{tab-Development}. To assess the models' sensitivity on different \textbf{\textit{Deployment-Gap Scenario}}, we conduct the experiments on the V2V CARLA Towns testing set of OPV2V dataset with the difference of \textit{Noisy Setting}. The  Fig.~\ref{fig:development-gap-reasult} depicts the higher robustness of our S2R-UViT against other fusion methods in the deployment gap.

\noindent \textbf{Performance in Sim2Real  Scenario.}
The 3D object detection result on the real-world V2V4Real testing set of \textbf{\textit{Sim2Real  Scenario}} is presented in Table~\ref{tab:domaingap}. Among all intermediate fusion methods without domain adaptation, our proposed S2R-UViT achieves the best performance by eliminating the deployment gap in the real V2VReal testing set. After applying domain adaptation methods \textit{e.g.}, GRL~\cite{ganin2015unsupervised} and AdvGRL~\cite{li2023domain}, all methods have improved performance. For example, V2X-ViT is improved by $12.1\%$/$5.3\%$ for AP@0.5/0.7 with GRL, and $13.0\%$/$5.4\%$ for AP@0.5/0.7 with AdvGRL. Our S2R-ViT with S2R-AFA achieves the $44.1\%$/$17.0\%$ for AP@0.5/0.7 as the best performance with improvement of $7.4\%$/$3.2\%$ than S2R-UViT. The examples of 3D object detection results on V2V4Real testing set under \textbf{\textit{Sim2Real Scenario}} are visualized in Fig.~\ref{fig:detect}, where our S2R-ViT generates more accurate 3D detection results.

\noindent \textbf{Ablation Study.}
As Table~\ref{tab:domaingap} depicts, 
all of the proposed components within S2R-ViT have contributed to enhanced detection performance. Specifically, our S2R-UViT model achieves the highest detection performance among all state-of-the-art V2V methods, which is $3.3\%$ and $0.8\%$ higher than the second-best performance model CoBEVT in AP@0.5 and AP@0.7 respectively. Adding our S2R-AFA, our proposed S2R-ViT achieves $44.1\%$ and $17.0\%$ in AP@0.5 and AP@0.7, with the improvement of $7.4\%$ in AP@0.5 and $3.2\%$ in AP@0.7.


\section{Conclusions}\label{Sec:Conclusions}

This paper is the first work that investigates the domain gap on multi-agent cooperation perception from simulation to reality, specifically focusing on the deployment gap and feature gap in point cloud-based 3D object detection. Based on the analysis, we present the first Simulation-to-Reality transfer learning framework using a novel Vision Transformer, named S2R-ViT, to mitigate these two types of domain gaps, which mainly contain an Uncertainty-aware Vision Transformer and an Agent-based Feature Adaptation module. The
experiment shows the effectiveness of S2R-ViT. 
This research presents a significant step forward in the multi-agent cooperation perception from  simulation to reality. 

\bibliographystyle{IEEEtran}
\bibliography{Jinlong}

\end{document}